\documentclass[runningheads]{llncs}
\usepackage{graphicx}
\usepackage{subfigure}
\usepackage{pdfpages}
\usepackage{amsmath,amssymb} 
\usepackage{color}
\usepackage{hyperref}

\makeatletter
\newcommand{\printfnsymbol}[1]{%
  \textsuperscript{\@fnsymbol{#1}}%
}
\makeatother

\begin{document}
\pagestyle{headings}
\mainmatter
\def\ECCV18SubNumber{2571}  

\title{Learning to Detect and Track Visible and Occluded Body Joints in a Virtual World} 
\titlerunning{Learning to Detect and Track}

\author{Matteo Fabbri\thanks{Equal contribution.}\and
Fabio Lanzi\printfnsymbol{1}\and
Simone Calderara\printfnsymbol{1}\and
Andrea Palazzi\and
Roberto Vezzani\and
Rita Cucchiara
}
%
\authorrunning{M. Fabbri et al.}
%

\institute{Department of Engineering ``Enzo Ferrari''\\
University of Modena and Reggio Emilia, Italy\\
\email{\{name.surname\}@unimore.it}}

\maketitle
\begin{abstract}
Multi-People Tracking in an open-world setting requires a special effort in precise detection. Moreover, temporal continuity in the detection phase gains more importance when scene cluttering introduces the challenging problems of occluded targets. For the purpose, we propose a deep network architecture that jointly extracts people body parts and associates them across short temporal spans. Our model explicitly deals with occluded body parts, by hallucinating plausible solutions of not visible joints. We propose a new end-to-end architecture composed by four branches (\textit{visible heatmaps}, \textit{occluded heatmaps}, \textit{part affinity fields} and \textit{temporal affinity fields}) fed by a \textit{time linker} feature extractor. To overcome the lack of surveillance data with tracking, body part and occlusion annotations we created the vastest Computer Graphics dataset for people tracking in urban scenarios by exploiting a photorealistic videogame. It is up to now the vastest dataset (about 500.000 frames, almost 10 million body poses) of human body parts for people tracking in urban scenarios. Our architecture trained on virtual data exhibits good generalization capabilities also on public real tracking benchmarks, when image resolution and sharpness are high enough, producing reliable tracklets useful for further batch data association or re-id modules.

\end{abstract}
\section{Introduction}
Multi-People Tracking (MPT) is one of the most established fields in computer vision. It has been recently fostered by the availability of comprehensive public benchmarks and data \cite{mot16,andriluka2018posetrack}. Often, MPT approaches have been casted in the \emph{tracking by detection paradigm} where a pedestrian detector extracts candidate objects and a further association mechanism arranges them in a temporally consistent trajectory \cite{solera,Rezatofighi_2015_ICCV,shahflow}. 
Nevertheless, in the last years several researchers \cite{zisserman2017,solera} raised the question on whether these two phases would be disentangled or considered two sides of the same problem. The strong influence between detection accuracy and tracking performance \cite{solera} suggests considering detection and tracking as two parts of a unique problem that should be addressed end-to-end at least for short-term setups. 
In this work, we advocate for an integrated approach between detection and short-term tracking that can serve as a proxy for more complex association method either belonging to the tracking or re-id family of techniques.
To this aim, we propose:
\begin{itemize}
\item an end-to-end deep network, called \textit{THOPA-net (Temporal Heatmaps and Occlusions based body Part Association)} that jointly locates people body parts and associates them across short temporal spans. This is achievable with modern deep learning architectures that exhibit terrific performance in body part location \cite{cao2017realtime} but, mostly, neglect the temporal contribution. 
For the purpose, we propose a bottom-up human pose estimation network with a temporal coherency module that jointly enhances the detection accuracy and allows for short-term tracking; 
\item an explicit method for dealing with occluded body parts that exploits the capability of deep networks of hallucinating feasible solutions;
\end{itemize}
Results are very encouraging in their precision also in crowded scenes.
Our experiments tell us that the problem is less dependent on the details or the realism of the shape than one could imagine; instead, it is more affected by the image quality and resolution that are extremely high in Computer Graphics (CG) generated datasets. 
Nevertheless, experiments on real MPT dataset \cite{mot16,andriluka2018posetrack} demonstrate that the model can transfer positively towards real scenarios.

\section{Related Works}
Human pose estimation in images has made important progress over the last few years \cite{carreira2016human,wei2016convolutional,hu2016bottom,newell2016stacked,bulat2016human}. However, those techniques assume only one person per image and are not suitable for videos of multiple people that occlude each other. The natural extension of single-person pose estimation, i.e, multi-person pose estimation, has therefore gained much importance recently being able of handling situations with a varying number of people \cite{pishchulin2016deepcut,insafutdinov2016deepercut,iqbal2016multi,papandreou2017towards,newell2017associative,cao2017realtime,levinkov2017joint}. Among them, \cite{papandreou2017towards} uses graph decomposition and node labeling with local search while \cite{newell2017associative} introduces associative embeddings to simultaneously generate and group body joints detections.  An end-to-end architecture for jointly learning body parts and their association is proposed by \cite{cao2017realtime} while  \cite{papandreou2017towards}, instead, exploits a two-stage approach, consisting of a person detection stage followed by a keypoint estimation for each person. Moreover, \cite{pishchulin2016deepcut,insafutdinov2016deepercut,iqbal2016multi} jointly estimate multiple poses in the image, while also handling truncations and occlusions. However, those methods still rely on a separate people detector and do not perform well in cluttered situations.\\
\begin{table}[t!]
\scriptsize
\begin{center}
\caption{Overview of the publicly available datasets for Pose Estimation and MPT in videos. For each dataset we reported the numbers of clips, annotated frames and people per frame, as well as the availability of 3D data, occlusion labels, tracking information, pose estimation annotations and data type}
\setlength{\tabcolsep}{4pt}
\label{table:datasets}
\begin{tabular}{lllccccccc}
\hline\noalign{\smallskip}
Dataset 			& \#Clips  & \#Frames 		& \#PpF & 3D 		& Occl.			& Tracking 		& Pose Est.	 	& Type 		 \\
\noalign{\smallskip}
\hline
\noalign{\smallskip}	
Penn Action \cite{zhang2013actemes}		& 2,326		&	 159,633 	& 1		&	 			&				& 				& \checkmark 	& sports \\
JHMDB \cite{jhuang2013towards}				& 5,100		&	 31,838 	& 1 	& 				&				& 				& \checkmark 	& diverse \\
YouTube Pose \cite{charles2016personalizing}		& 50 		&	 5,000 		& 1 	& 				&				& 				& \checkmark 	& diverse \\
Video Pose 2.0 	\cite{sapp2011parsing}	& 44		&	 1,286 		& 1 	& 				&				& 				& \checkmark 	& diverse \\
\noalign{\smallskip}
\hline
\noalign{\smallskip}
Posetrack \cite{andriluka2018posetrack}			& 514 		& 23,000 		& 1-13 	&  		  		&				& \checkmark 	& \checkmark 	& diverse \\
MOT-16 	\cite{mot16}			& 14 		& 11,235 		& 6-51 	&  				&	\checkmark	& \checkmark 	& 				& urban  \\
\noalign{\smallskip}
\hline
\noalign{\smallskip}
JTA 						& 512 		& 460,800		& 0-60 	& \checkmark	& \checkmark 	& \checkmark 	& \checkmark 	& urban  \\
\hline
\end{tabular}
\end{center}
\end{table}

Single person pose estimation in videos has been addressed by several researchers, \cite{jain2014modeep,zhang2015human,pfister2015flowing,gkioxari2016chained}.   
Nevertheless, all those methods improve the pose estimation accuracy by exploiting temporal smoothing constraints or optical flow data, but neglect the case of multiple overlapping people.\\
In recent years, online tracking has been successfully extended to scenarios with multiple targets \cite{poi,sort,savareseiccv,chuiccv,baetpami,cavallaro}. In contrast to single target tracking approaches, which rely on sophisticated appearance models to track a single entity in subsequent frames, multiple target tracking does not rely solely on appearance models. \cite{poi} exploits a high-performance detector with a deep learning appearance feature while \cite{savareseiccv} presents an online method that encodes long-term temporal dependencies across multiple cues. \cite{chuiccv}, on the other hand, introduces spatial-temporal attention mechanism to handle the drift caused by occlusion and interaction among targets. \cite{baetpami} solves the online multi-object tracking problem by associating tracklets and detections in different ways according to their confidence values and \cite{cavallaro} exploits both high and low confidence target detections in a probability hypothesis density particle filter framework.\\
In this work, we address the problem of multi-person pose estimation in videos jointly with the goal of multiple people tracking. Early works that approach the problem \cite{andriluka2008people,izadinia20122t} do not tackle pose estimation and tracking simultaneously, but rather target on multi-person tracking alone. More recent methods \cite{iqbal2017posetrack,insafutdinov2017arttrack}, which rely on graph partitioning approaches closely related to \cite{pishchulin2016deepcut,insafutdinov2016deepercut,iqbal2016multi}, simultaneously estimate the pose of multiple people and track them over time but do not cope with urban scenarios that are dominated by targets occlusions, scene clutterness and scale variations. In contrast to \cite{iqbal2017posetrack,insafutdinov2017arttrack} we do not tackle the problem as a graph partitioning approach. Instead, we aim at simplifying the tracking problem by providing accurate detections robust to occlusions by reasoning directly at video level.\\
The most widely used publicly available datasets for human pose estimation in videos are presented in Tab. \ref{table:datasets}. \cite{zhang2013actemes,jhuang2013towards,charles2016personalizing} provide annotations for the single-person subtask of person pose estimation. Only Posetrack \cite{andriluka2018posetrack} has a multi-person perspective with tracking annotations but not provide them in the surveillance context.
The reference benchmark for evaluation of multi-person tracking is \cite{mot16} which provides challenging sequences of crowded urban scenes with severe occlusions and scale variations. However, it pursuits no pose estimation task and only provides bounding boxes as annotations.
Our virtual world dataset instead, aim at taking the best of both worlds by merging precise pose and tracking annotations in realistic urban scenarios. This is indeed feasible when the ground truth can be automatically computed exploiting highly photorealistic CG environments.

\section{JTA Dataset}
\begin{figure}[b!]
\centering
\includegraphics[width=1.0\columnwidth]{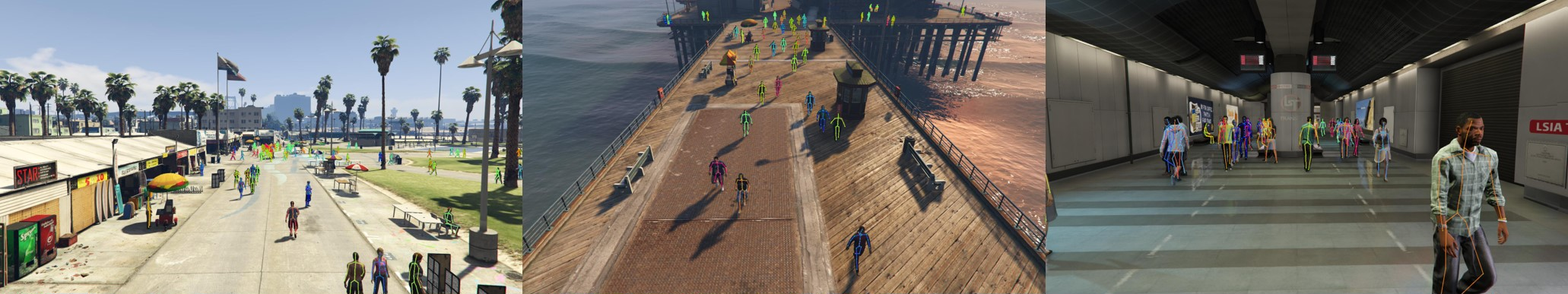}
\caption{Examples from the JTA dataset exhibiting its variety in viewpoints, number of people and scenarios. Ground truth joints are superimposed to the original images. See supplementary material for further examples}
\label{fig:jta}
\end{figure}

We collected a massive dataset JTA (Joint Track Auto) for pedestrian pose estimation and tracking in urban scenarios by exploiting the highly photorealistic video game \textit{Grand Theft Auto V} developed by \textit{Rockstar North}.
The collected videos feature a vast number of different body poses, in several urban scenarios at varying illumination conditions and viewpoints, Figure \ref{fig:jta}.
Moreover, every clip comes with a precise annotation of visible and occluded body parts, people tracking with 2D and 3D coordinates in the game virtual world. In terms of completeness, our JTA dataset overcomes all the limitation of existing dataset in terms of number of entities and available annotations, Table \ref{table:datasets}.  
In order to virtually re-create real-world scenarios we manually directed the scenes by developing a game modification that interacts synchronously with the video game's engine. The developed module allowed us to generate and record natural pedestrian flows recreating people behaviors specific to the most crowded areas. Moreover, exploiting the game's APIs, the software can handle people actions: in clips, people occasionally perform natural actions like sitting, running, chatting, talking on the phone, drinking or smoking.
Each video contains a number of people ranging between 0 and 60 with an average of more than 21 people, totaling almost 10M annotated body poses over 460,800 densely annotated frames. The distance from the camera ranges between 0.1 and 100 meters, resulting in pedestrian heights between 20 and 1100 pixels (see supplementary material for further details).
\setlength{\tabcolsep}{1.4pt}
We collected a set of 512 Full HD videos, 30 seconds long, recorded at 30 fps.
We halve the sequences into 256 videos for training and 256 for testing purposes.
Through the game modification, we access the game renderer for automatically annotating the same 14 body parts in \cite{mpii} and \cite{andriluka2018posetrack} in order to foster cross-dataset experiments.
In each video, we assigned a unique identifier to every pedestrian that appears in the scene. The identifier remains the same throughout the entire video even if the pedestrian moves out the field-of-view. This feature could foster person re-identification research despite not being the target of this work.
Our dataset also provides \emph{occlusion} and \emph{self-occlusion} flags. Each joint is marked as occluded if it is not directly visible from the camera point of view and it is occluded by objects or other pedestrians. Instead, a joint is marked as self-occluded if it is occluded by the same person to whom the joint belongs. As for joints annotation, occlusion annotation is captured by accessing the game renderer.
JTA Dataset also provides accurate 3D information: for each annotated joint, as well as having the 2D coordinates of the location in the image, we also provide the 3D coordinates of the location in the simulator's space. Differently from Posetrack \cite{andriluka2018posetrack}, which uses the annotated head bounding boxes as an estimation of the absolute scale of the person, we provide the precise scale of each pedestrian through the 3D annotation. The dataset, along with the game modification, are freely accessible\footnote{\url{http://imagelab.ing.unimore.it/jta}}.
\section{THOPA-net}
Our approach exploits both intra-frame and inter-frame information in order to jointly solve the problem of multi-person pose estimation and tracking in videos. For individual frames, we extended the architecture in \cite{cao2017realtime} by integrating a branch for handling occluded joints in the detection process. 
Subsequently, we propose a temporal linking network to integrate temporal consistency in the process and jointly achieve detection and short-term tracking. 
The Single Image model, Figure \ref{fig:sigleimage}, takes an RGB frame of size $w \times h$ as input and produces, as output, the pose prediction for every person in the image. 
Conversely, the complete architecture, Figure \ref{fig:time}, takes a clip of $N$ frames as input and outputs the pose prediction for the last frame of the clip and the temporal links with the previous frame.
\subsection{Single Image Pose Prediction}
\label{subsec:single_image}
\begin{figure}[t!]
\centering
\includegraphics[width=0.9\linewidth]{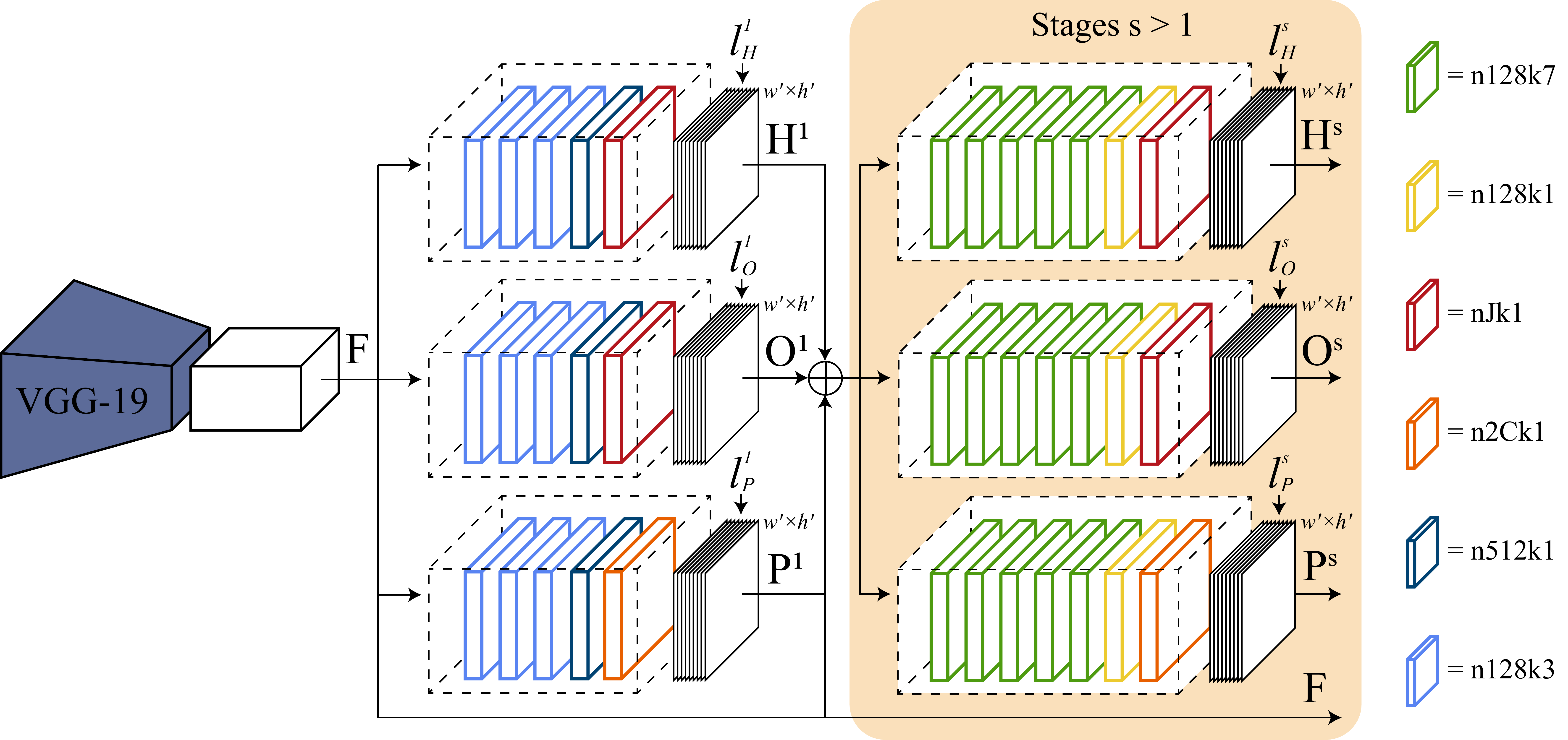} 
\caption{Architecture of the three-branch multi-stage CNN with corresponding kernel size (k) and number of feature maps (n) indicated for each convolutional layer}
\label{fig:sigleimage}
\end{figure}

Our single image model, Figure \ref{fig:sigleimage}, consists of an initial feature extractor based on the first 10 layers of VGG-19 \cite{simonyan2014very} pretrained on COCO 2016 keypoints dataset \cite{lin2014microsoft}. The computed feature maps are subsequently processed by a three-branch multi-stage CNN where each branch focuses on a different aspect of body pose estimation: the first branch predicts the heatmaps of the visible parts, the second branch predicts the heatmaps of the occluded parts and the third branch predicts the part affinity fields (PAFs), which are vector fields used to link parts together.
Note that, oppositely to \cite{cao2017realtime}, we employed a different branch for the occlusion detection task. It is straightforward that visible and occluded body parts detection are two related but distinct tasks. The features used by the network in order to detect the location of a body part are different from those needed to estimate the location of an occluded one. Nevertheless, the two problems are entangled together since visible parts allow to estimate the missing ones.
In fact, the network exploits contextual cues in order to perform the desired prediction, and the presence of a joint is indeed strongly influenced by the person's silhouette (e.g. a foot detection mechanism relies heavily on the presence of a leg, thus a visible foot detection may trigger even though the foot is not completely visible). 
Each branch is, in turn, an iterative predictor that refines the predictions at each subsequent stage applying intermediate supervision in order to address the vanishing gradient problem. Apart from the first stage, which takes as input only the features provided by VGG-19, the consecutive stages integrate the same features with the predictions from the branches at the previous stage. Consequently, information flow across the different branches and in particular both visible and occluded joints detection are entangled in the process.\\  
We apply, for each branch, a different loss function at the end of each stage. The loss is a SSE loss between estimated predictions and ground truth, masked by a mask $M$ in order to
not penalize occluded joints in the visible branch. Specifically, for the generic output of each branch $X^s$ of stage $s \in \{1,\dots,S\}$ and the ground truth $X^*$ we have the loss function:
\begin{align}
 l_X^s = \sum\limits_i \ \ \sum_{x=1}^{w^\prime} \sum_{y=1}^{h^\prime} M(x,y) \odot (X_i^s(x,y) - X_i^{\ast}(x,y))^2,
\label{eq:loss_single}
 \end{align}
where $X$ is in turn $H$ for visible joints heatmaps, $O$ for occluded ones and $P$ for affinity fields; the outer summation spans the $J$ number of joints for $H$ and $O$ and the $C$ number of limbs for $P$. $H^s$, $O^{s}$ and $P^s$ sizes $(w^\prime,h^\prime)$ are eight times smaller than the input 
due to VGG19 max pooling operations.
Eventually, the overall objective becomes $L = \sum_{s=1}^S (l_H^s + l_O^s + l_P^s)$.
\subsection{Temporal Consistency Branch}
\begin{figure}[t!]
\centering
\includegraphics[width=0.84\linewidth]{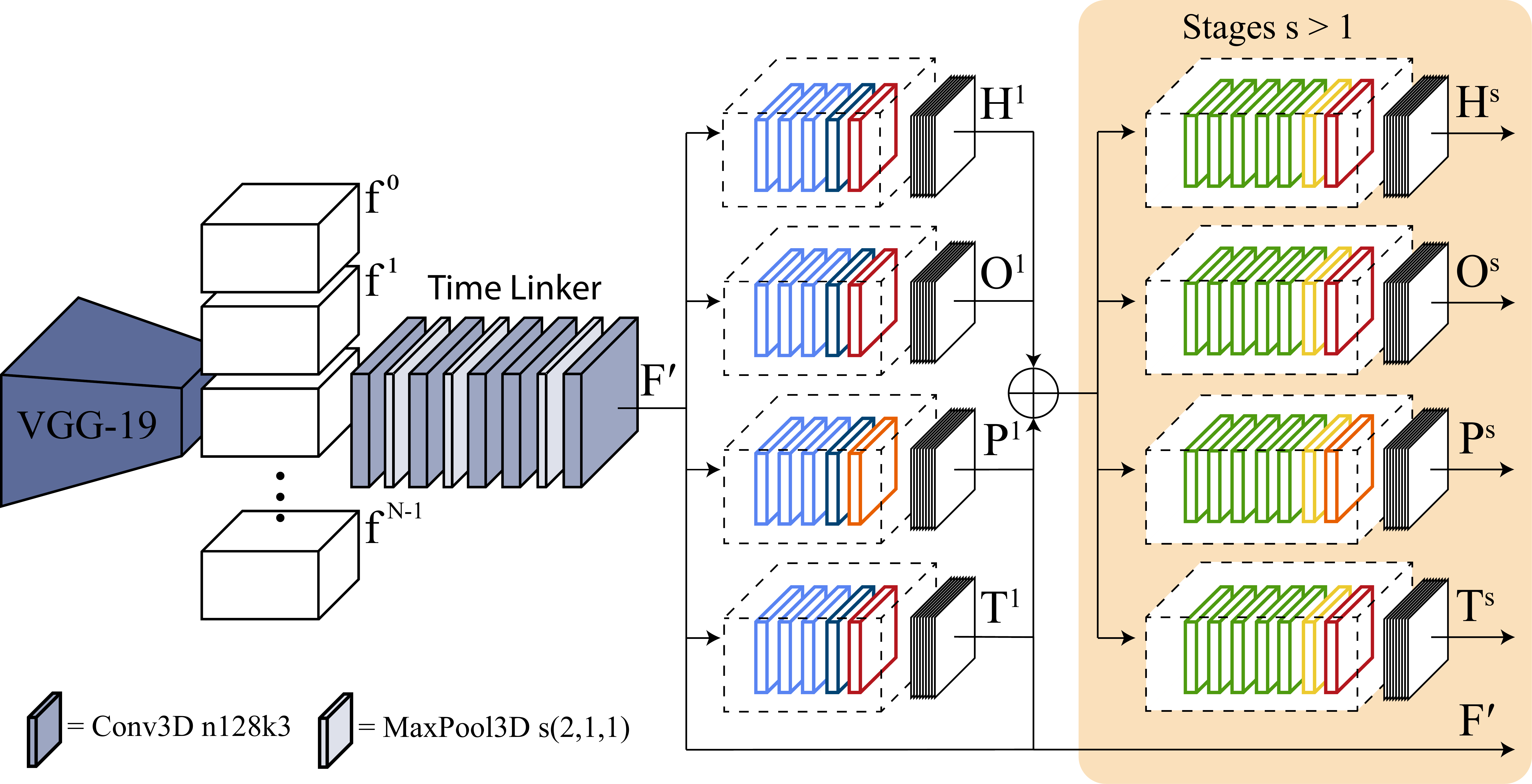}
\caption{Architecture of our method that encompass pose estimation and tracking in an end-to-end fashion. The MaxPool3D perform pooling operations only in the temporal dimension with stride s}
\label{fig:time}
\end{figure}

In order to jointly solve the problem of multi-person pose estimation and tracking we enhance the Single Image model by adding our novel temporal network, Figure \ref{fig:time}. The temporal model takes as input $N$ RGB frames of size $w \times h$ and produces, as output, the temporal affinity fields (TAFs), as well as heatmaps and part affinity fields.
TAFs, like PAFs, are vector fields that link body parts but, oppositely to PAFs, are focused on temporal links instead of spatial ones.
In detail, PAFs connect different types of body parts intra-frame while TAFs, instead, connect the same types of body parts inter-frame, e.g, they connect heads belonging to the same person in two subsequent frames. The TAF field is, in fact, a proxy of the motion of the body parts and provide the expected location of the same body part in the previous frame and can be used both for boosting the body parts detection and for associating body parts detections in time.
At a given time $t_0$, our architecture takes frames $I^t \in \mathbb{R}^{w\times h \times 3}$ with $t \in \{t_0, t_{-\tau}, t_{-2\tau}, \dots, t_{-N\tau+1} \}$ and pushes them through the VGG19 feature extractor, described in Section \ref{subsec:single_image}, to obtain $N$ feature tensors $f^t \in \mathbb{R}^{w^\prime\times h^\prime \times r}$ where $r$ is the number of channels of the feature tensor. Those tensors are then concatenated over the temporal dimension obtaining $F \in \mathbb{R}^{w^\prime\times h^\prime \times r\times N}$. $F$ is consecutively fed to a cascade of 3D convolution blocks that, in turn, capture the temporal patterns of the body part features and distill them by temporal max pooling until we achieve a feature tensor $F^\prime \in \mathbb{R}^{w^\prime\times h^\prime \times r}$, Figure \ref{fig:time}.
As in Section \ref{subsec:single_image}, the feature maps are passed through a multi-branch multi-stage CNN. 
Moreover, we add to the Single Image architecture a fourth branch for handling the TAFs prediction. As a consequence, after the first stage, temporal information flow to all the branches of the network and acts as a prior for body part estimation (visible and occluded) and PAFs computation. 
The complete network objective function then becomes $L = \sum_{s=1}^S (l_H^s + l_O^s + l_P^s + l_T^s)$
where
\begin{align}
l_T^s = \sum_{j=1}^J \ \ \sum_{x=1}^{w^\prime} \sum_{y=1}^{h^\prime} M(x,y) \odot (T_j^s(x,y) - T_j^{\ast}(x,y))^2 
\end{align}
is the loss function computed between the ground truth $T_j^{\ast}$ and the prediction $T_j^{s}$ at each stage $s$. The set $T = (T_1, T_2, \dots, T_j)$ has $J$ vector fields, one for each part, with $T_j \in \mathbb{R}^{w\times h}, j \in \{1,\dots,J\}$.
\subsection{Training Procedure}
\label{sec:training}
\begin{figure}
\centering
\subfigure[]{\includegraphics[width=0.35\linewidth]{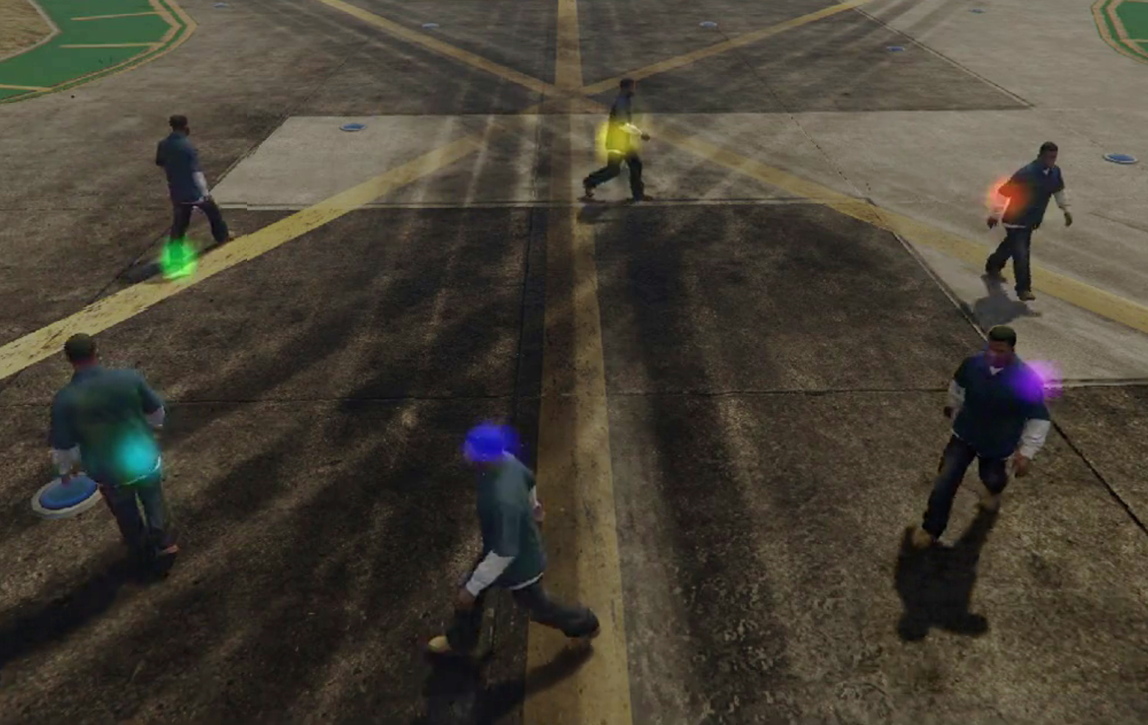}} 
\subfigure[]{\includegraphics[width=0.35\linewidth]{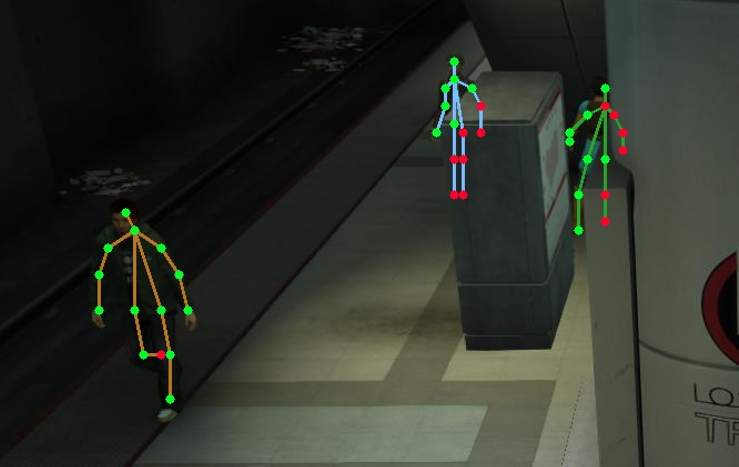}} 
\caption{(a) Visualization of TAFs for different parts: for clarity, we show a single joint TAF for each person where color encodes direction. (b) Pose prediction performed on JTA dataset which distinguish between visible and occluded joints}
\label{fig:examples}
\end{figure}

During training, we generate both the ground truth heatmaps $H^\ast$ and $O^\ast$ from the annotated keypoint coordinates by placing at the keypoint location a 2D Gaussian with its variance conditioned by the true metric distance, $d$, of the keypoint from the camera. Oppositely to \cite{cao2017realtime}, by smoothing the Gaussian using distances, it is possible to achieve heatmaps of different sizes proportional to the scale of the person itself. This process is of particular importance to force scale awareness in the network and avoiding the need of multi scale branches.
For example, given a visible heatmap $H_j$, let $q_{j,k} \in \mathbb{R}^2$ be the ground truth location of the body part $j$ of the person $k$. For each body part $j$ the ground truth $H_j^\ast$ at location $p \in \mathbb{R}^2$ results:
\begin{align}
H_j^\ast(p) = \max_k \exp{\bigg(-\frac{\big\lVert p-q_{j,k} \big\rVert_2^2}{\sigma^2}\bigg)}, \qquad \sigma = \exp{\bigg(1-\frac{d}{\alpha}\bigg)}
\end{align}
where $\sigma$ regulates the spread of the peak in function of the distance $d$ of each joint from the camera. In our experiments we choose $\alpha$ equals to $20$. \\
Instead, each location $p$ of ground truth part affinity fields $P_{c,k}^\ast$ is equal to the unit vector (with the same direction of the limb) if the point $p$ belongs to the limb. The points belonging to the limb are those within a distance threshold of the line segment that connect the pair of body parts.
For each frame, the ground truth part affinity fields are the two channels image containing the average of the PAFs of all people.
As previously stated, by extending the concept of PAFs to the temporal dimension, we propose the novel TAFs representation which encodes short-term tubes of body parts across multiple frames (as shown in Figure \ref{fig:examples}.(b)). The temporal affinity field is a 2D vector field, for each body part, that points to the location of the same body part in the previous frame.
Consider a body part $j$ of a person $k$ at frame $t$ and let $q_{j,k}^{t-1}$ and $q_{j,k}^{t}$ be their ground truth positions at frame $t-1$ and $t$ respectively. If a point $p$ lies on the path crossed by the body part $j$ between $t-1$ and $t$, the value at $T_{j,k}^\ast(p)$ is a unit vector pointing from $j$ at time $t$ to $j$ at time $t-1$; for all other points the vector is zero. We computed ground truth TAFs using the same strategy exploited for PAFs.

\subsection{Spatio-Temporal Multi-Person Joints Association} 
In order to connect body parts into skeletons we take into account two different contributions both at frame level (PAF) and at temporal level (TAF). First, the joints heatmaps are non-maxima suppressed to obtain a set of discrete locations, $D_j$, for multiple people, where $D_j = \{d_j^m:$ for $j \in \{1, \dots, J\}, m \in \{1,\dots, N_j \}\}$ and $N_j$ is the number of candidates of part $j$, and $J$ the number of joint types.\\
We associate joints by defining a variable $z_{j_1 j_2}^{mn} \in \{0,1\}$ to indicate whether two joints candidates $d_{j_1}^m$ and $d_{j_2}^n$ are connected.
Consequently, the objective is to find the optimal assignment for the set of possible connections, $Z = \{ z_{j_1 j_2}^{mn} : $ for $j_1,j_2 \in \{1,\dots, J\}, m \in \{1, \dots, N_{j_1} \}, n \in \{1, \dots, N_{j_2} \} \}$.
To this aim we score every candidate limb (i.e. a pair of joints) spatially and temporally by computing the line integral along PAFs, $E$ and TAFs, $G$:  
\begin{align}
E(d_{j_1}, d_{j_2}) = \int_{u=0}^{u=1} PAF(p(u)) \cdot \frac{d_{j_2} - d_{j_1}}{\big\lVert d_{j_2} - d_{j_1} \big\rVert_2} du
\end{align}
\begin{align}
G(d_j, \hat{d}_j) = \int_{u=0}^{u=1} TAF(t(u)) \cdot \frac{\hat{d}_{j} - d_{j}}{\big\lVert \hat{d}_{j} - d_{j} \big\rVert_2} du
\end{align}
where $p(u)$ linearly interpolates the locations along the line connecting two joints $d_{j_2}$ and $d_{j_1}$ and $t(u)$ acts analogously for two joints $\hat{d}_{j}$ at frame $t-1$ and $d_{j}$ at frame $t$.\\ 
We then maximize the overall association score $E_c$ for limb type $c$ and every subset of allowed connection $Z_c$ (i.e. anatomically plausible connections):
\begin{align}
\label{eq:max}
\max_{Z_c} E_c = \max_{Z_c} \sum_{m\in D_{j_1}}\sum_{n\in D_{j_2}} (E(d_{j_1}^m, d_{j_2}^n) + \alpha E(\hat{d}_{j_1}^m, \hat{d}_{j_2}^n))\cdot z_{j_1 j_2}^{mn},
\end{align}
subject to $\sum_{n\in D_{j_2}}  z_{j_1 j_2}^{mn} \leq 1, \forall m \in D_{j_1}$ and $\sum_{n\in D_{j_2}}  z_{j_1 j_2}^{mn} \leq 1, \forall m \in D_{j_1}$ where
\begin{align}
\hat{d}_{j_1}^m = \arg \max_{\hat{d}_{j_1}^b} G(d_{j_1}^m, \hat{d}_{j_1}^b), \qquad \qquad
\hat{d}_{j_2}^n = \arg \max_{\hat{d}_{j_2}^q} G(d_{j_2}^n, \hat{d}_{j_2}^q)
\end{align}
are the joints at frame $t-1$ that maximize the temporal consistency along the TAF where $b$ and $q$ span the indexes of the people detected at the previous frame.\\
In principle, Equation \eqref{eq:max} mixes both the contribution coming from the PAF in the current frame and the contribution coming from the PAF obtained by warping, in the previous frame, the candidate joints along the best TAF lines.\\
In order to speed up the computation, we maximize iteratively Equation \eqref{eq:max} by considering only the subsets of joints inside a radius at twice the size of the skeletons in the previous frame at the same location. 
The complete skeletons are then built, by maximizing, for the limbs type set $C$, $E = \sum_{c=1}^C \max_{Z_c} E_c$.
\begin{figure}[t!]
\centering
\includegraphics[width=0.9\linewidth]{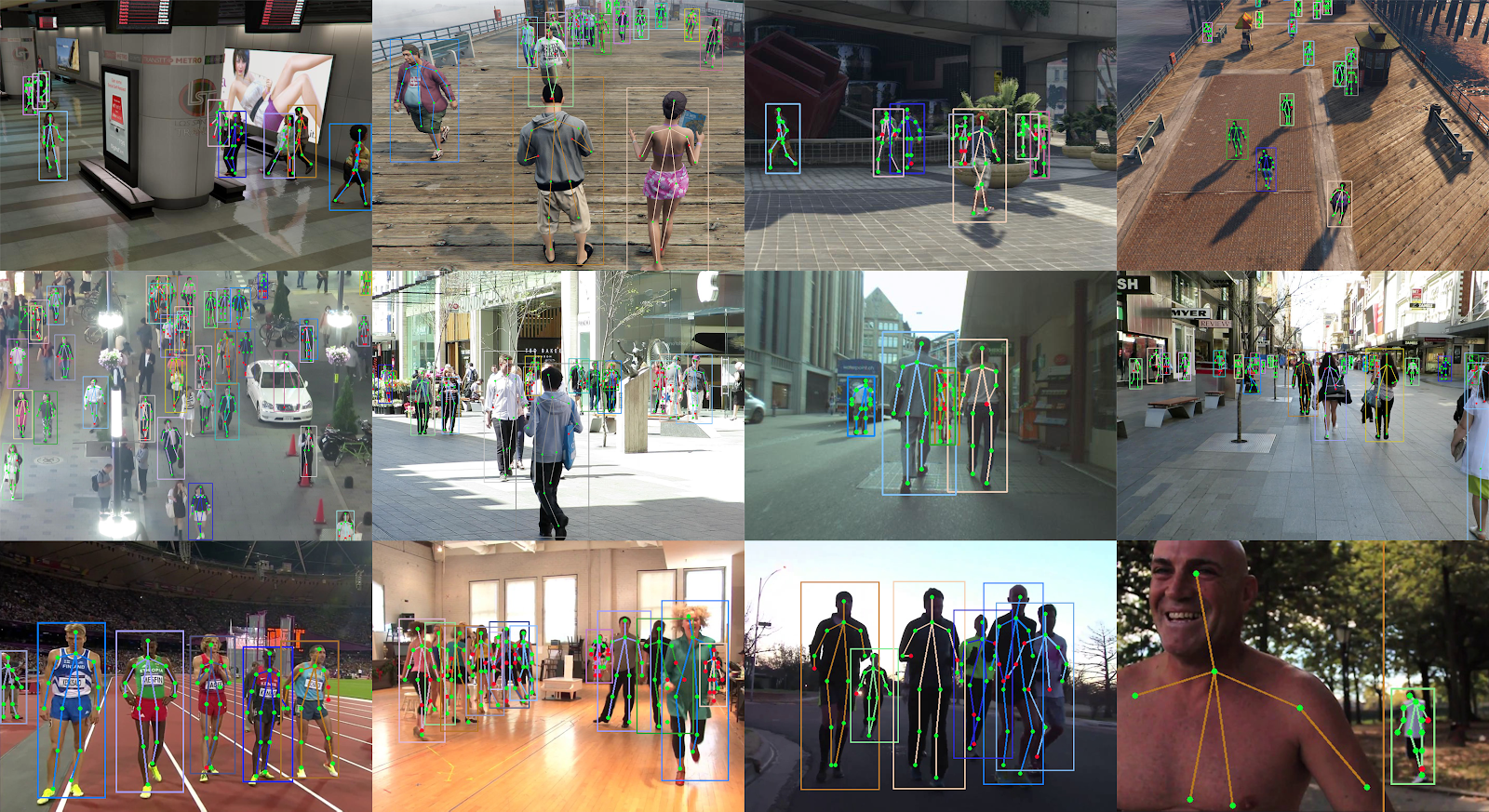}
\caption{Qualitative results of THOPA-net on JTA (top row), MOT-16 (middle row) and PoseTrack (bottom row)} 
\label{fig:results}
\end{figure}
\section{Experiments}
We conducted experiments in two different contexts, either on our virtual world dataset JTA and on real data.
In the virtual world scenario, we evaluated the capability of the proposed architecture of both reliably extracting people joints and successfully associating them along the temporal dimension. Real data experiments instead, aimed at empirically demonstrating that our virtual world dataset can function as a good proxy for training deep models and to which extent it is necessary to fine-tune the network on real data. In fact, we purposely conducted the experiments either without retraining the network and testing it out-of-the-box or by fine-tuning the network on real data. Moreover, all the tracking experiments do not explicitly model the target appearance, but visual appearance is only taken into account when extracting TAFs, thus exploited only for very short-term target association (namely tracklet construction).
\begin{table}[t!]
\setlength{\tabcolsep}{5pt}
\centering
\caption{Detection results on JTA Dataset}
\label{tab:JTA}
\begin{tabular}{l|l||l|l|l}
\multicolumn{1}{c}{} & \multicolumn{1}{c}{\textbf{Joints} }                & \multicolumn{3}{c}{\textbf{Detection}  }                                                           \\
\multicolumn{1}{c}{} & \multicolumn{1}{c}{Mean Average Prec.} & \multicolumn{1}{c}{Precision} & \multicolumn{1}{c}{Recall} & \multicolumn{1}{c}{F1 Score} \\
\hline
Single Image no occ  &      50.9     &      81.5        &      64.1  &      71.6      \\
Single Image + occ   &      56.3     &      87.9        &      71.8  &      78.4      \\
Complete                 &     \textbf{59.3}     &     \textbf{92.1}        &     \textbf{77.4}  &     \textbf{83.9}   \\
\hline
\cite{cao2017realtime}   &      50.1     &      86.3        &     55.8  &      69.5    
\end{tabular}
\end{table}
\subsection{Experiments on JTA}
We tested our proposal on our virtual world scenario in order to evaluate both the joints extraction accuracy and the tracking capabilities. We started from the pre-trained VGG19 weights as the feature extractor and we trained our model end-to-end allowing features fine-tuning. 
For the temporal branch we randomly split every sequence into 1 second long clips. Subsequently, we uniformly subsampled every clip obtaining 8 frames that are inputted to the temporal branch.
The train was performed by using ADAM optimizer with a learning rate of $10^{-4}$ and batch size equal to 16. We purposely kept the batch size relatively small because every frame carries a high number of different joints at different scales and locations leading to a reliable average gradient for the task.
\paragraph{Detection experiment}
We first performed a detection experiment in order to quantify the contribution of the individual branch of our architecture.
The detection experiment evaluated the location of people joints and the overall bounding box accuracy in terms of detection metrics.
Analogously to \cite{iqbal2017posetrack}, we used the PCKh (head-normalized probability of correct keypoint) metric, which considers a body joint to be correctly localized if the predicted location of the joint is within a certain threshold from the true location. 
Table \ref{tab:JTA} reports the results in term of mean average precision of joints location and bounding box detection metrics such as precision, recall and F1-score with an intersection over union threshold of $ 50\% $. 
We additionally ablated different branch of our architecture in order to empirically measure the contribution of every individual branch (i.e. the occlusion branch and the temporal branch).
By observing the Table we can confirm that the network benefits from the presence of the occlusion estimation branch both in terms of joints location accuracy and detection performances. This is due to two different positive effects given by occluded joints. The first is the chance of estimate/guess the position of a person even if visually strong occluded, the second is about maximizing the presence of body joints that greatly simplifies their clustering into skeletons and consequently the detection metrics results improved, Figure \ref{fig:examples}.(b).
Moreover, the temporal branch strengthens this process by adding short-term temporal consistency to the joints location. In fact, results indicate this boosts the performance leading to a more accurate joints detection in presence of people that overlaps in the scene. The improvement is due to the TAFs contribution that helps to disambiguate the association among body joints on the basis of the target direction, Figure \ref{fig:examples}.(a).
Additionally we compared with \cite{cao2017realtime} that was retrained on JTA and tested at 2 different scales (since the method does not deal with multiple scales), against which we score positively. The architecture in \cite{cao2017realtime} is the same as our \emph{Single Image no occ} model in Table \ref{tab:JTA}, with the only difference that the latter has been trained with distance rescaled versions of heatmaps and PAFs, according to Section \ref{sec:training}, and it deals with multiple scales without any input rescaling operation.

\paragraph{Tracking Experiment}
We additionally tested the extent of disentanglement between temporal short-term detection and people tracking by performing a complete tracking experiments on the JTA test set. The experiments have been carried out by processing 1 second clips with a stride of 1 frame and associating targets using a local nearest neighbour approach maximizing the TAFs scores. As previously introduced, the purpose of the experiment was to empirically validate the claim that mixing short-term tracking and detection can still provide acceptable overall tracking performance even when adopting a simple association frame-by-frame method. Secondly, this is indeed more evident when the association algorithm exploits more than a single control point (e.g. usually the bounding box lower midpoint), which is the case of tracking sets of joints. For the purpose, we compared against a hungarian based baseline (acting on the lower midpoint of the bounding box), \cite{solera}, inputed with either our detections and DPM \cite{dpm} ones.
Table \ref{tab:tracking_jta} reports results in terms of Clear MOT tracking metrics \cite{mot16}.
Results indicate that the network trained on the virtual world scores positively in terms of tracked entities but suffers of a high number of IDs and FRAGS. This behavior is motivated by the absence of a strong appearance model capable of re-associating the targets after long occlusions. Additionally, the motion model is purposely simple suggesting that a batch tracklet association procedure can lead to longer tracks and reduce switches and fragmentations.  
\begin{table}[t!]
\setlength{\tabcolsep}{5pt}
\centering
\caption{Tracking Results on JTA Dataset}
\label{tab:tracking_jta}
\begin{tabular}{l c|c|c|c|c|c|c|c}
                & MOTA & IDF1 & MT   & ML   & FP    & FN     & IDs & FRAG \\
\hline
\cite{solera} + our det & 57.4 & 57.3 & 45.3 & 21.7 & 40096 & 103831 & 15236 & 15569 \\
\cite{solera} + DPM det & 31.5 & 27.6 & 25.3 & 41.7 & 80096 & 170662 & 10575 & 19069 \\
THOPA-net & \textbf{59.3} & \textbf{63.2} & \textbf{48.1} & \textbf{19.4} & \textbf{40096} & \textbf{103662} & \textbf{10214} & \textbf{15211}    
\end{tabular}
\end{table}
\subsection{Tracking people in real data}

We tested our solution on real data with the purpose of evaluating the generalization capabilities of our model and its effectiveness in real surveillance scenarios. We choose to adopt two datasets: the commonly used MOT-16 Challenge Benchmark \cite{mot16} and the new PoseTrack Dataset \cite{andriluka2018posetrack}.

\paragraph{MOT-16.}
The MOT-16 Challenge Benchmark consists of 7 sequences in urban areas with varying resolution from 1980 $\times$ 1024 to 640 $\times$ 480 for a total number of approx 5000 frames and 3.5 minutes length. 
The benchmark exhibits strong challenges in terms of viewpoint changes, from top-mounted surveillance cameras to street level ones, Figure \ref{fig:results}.
All results are expressed in terms of Clear MOT metrics according to the benchmark protocol \cite{mot16} and as for the virtual world tracking experiment the tracks were associated by maximizing the TAF scores between detections.  
The network was end-to-end fine-tuned, with the exception of the occlusion branch. Fine-tuning was performed by considering the ground truth detections and inserting a default skeleton when our Single Image model scored a false negative obtaining an automatically annotated dataset. 

\begin{table}[t!]
\setlength{\tabcolsep}{5pt}
\centering
\caption{Results on MOT-16 benchmark ranked by MOTA score}
\label{tab:tracking_mot}
\begin{tabular}{l c|c|c|c|c|c|c|c}
                & MOTA & IDF1 & MT   & ML   & FP    & FN     & IDs & FRAG \\
\hline
\cite{poi}          & \textbf{66.1} & \textbf{65.1} & \textbf{34.0} & 20.8 & 5061  & \textbf{55914}  & 805  & 3093 \\
\cite{sort}         & 61.4 & 62.2 & 32.8 & \textbf{18.2} & 12852 & 56668  & 781  & 2008 \\
\textbf{THOPA-net}  & 56.0 & 29.2 & 25.2 & 27.9 & 9182  & 67059  & 4064 & 5557 \\   
\cite{savareseiccv} & 47.2 & 46.3 & 14.0 & 41.6 & \textbf{2681}  & 92856  & 774  & 1675 \\
\cite{chuiccv}      & 46.0 & 50.0 & 14.6 & 43.6 & 6895  & 91117  & \textbf{473}  & \textbf{1422} \\
\cite{baetpami}     & 43.9 & 45.1 & 10.7 & 44.4 & 6450  & 95175  & 676  & 1795 \\
\cite{cavallaro}    & 38.8 & 42.4 & 7.9  & 49.1 & 8114  & 102452 & 965  & 1657 \\
\end{tabular}
\end{table}

Table \ref{tab:tracking_mot} reports the results of our fine-tuned network compared with the best published state of the art competitors up to now. We include in the Table only online trackers, that are referred on the benchmark website as causal methods. The motivation is that our method performs tracking at low level, using TAFs, for framewise temporal association thus it configures as an online tracker. Additionally, it is always possible to consider our tracklets as an intermediate output and perform a subsequent global association by possibly assessing additional high level information such as strong appearance cues and re-id techniques.
Our method performs positively in terms of MOTA placing at the top positions. We observe a high IDS value and FRAG given by the fact that our output is an intermediate step between detections and long-term tracking.
Nevertheless, we remark that we purposely choose a trivial association method that does not force any strong continuity in terms of target trajectories, instead, we argue that given temporal consistency to the target detections the association among them results satisfying for short-term tracking applications. This is possible also thanks to the fact that we use several control points for association (i.e. the joints) that are in fact reliable cues when objects are close each other and the scene is cluttered.
Contrary to \cite{poi} and \cite{sort} our model do not employ strong appearance cues for re-identification. This suggests that the performance can be further improved by plugging a re-id module that connects tracks when targets are lost. 
Moreover, contrary to \cite{savareseiccv} we do not employ complex recurrent architecture to encode long-term dynamics. Nevertheless, the performances are comparable suggesting that when a tracker disposes of a plausible target candidate, even if occluded, the association simplify to keep subsequent frames temporally consistent that is indeed what our TAF branch do. Figure \ref{fig:results} shows qualitative results of our proposal.

\begin{figure}[t!]
\centering
\includegraphics[width=0.9\linewidth]{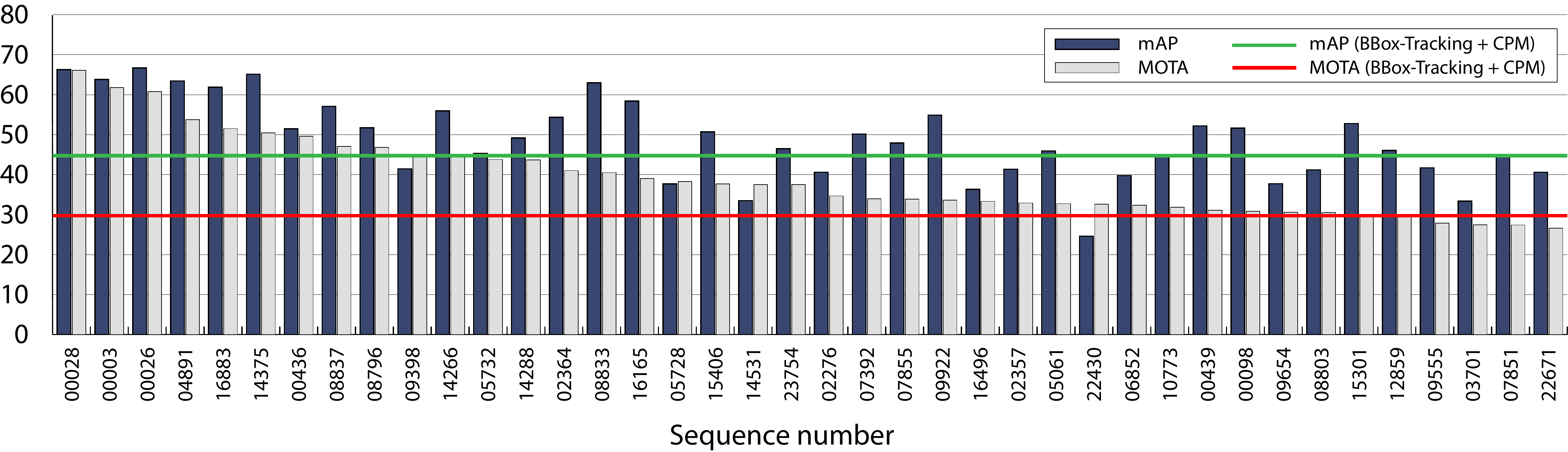}
\caption{Results on PoseTrack dataset compared with a BBox-Tracking + CPM (trained on MPII) baseline (used also in \cite{iqbal2017posetrack}; red/green lines are the average of performances on the selected sequences to avoid plot clutter)}
\label{fig:bar_plt2}
\end{figure}

\paragraph{PoseTrack.}
The PoseTrack Dataset is a large-scale benchmark for multi-person pose estimation
and tracking in videos. It contains 550 videos including around 23,000 annotated frames, split into 292, 50, 208 videos for training, validation and testing, respectively. The annotations include 15 body keypoints location, a unique person id and a head bounding box for each person instance.
We tested our solution on a subset of PoseTrack Dataset with surveillance like features (e.g. people standing, walking, etc.). We remark that PoseTack exhibits different features w.r.t. surveillance context in which the targets number is higher and the camera FoV is mostly a far FoV. In Fig. \ref{fig:bar_plt2} we show MOTA and mAP results of THOPA-net on PoseTrack sequences (solely using synthetic data for training). We used training and validation sequences in order to obtain per-sequence results.
The results are satisfying (see Fig \ref{fig:results}) even if the network is trained solely on CG data suggesting it could be a viable solution for fostering research in the joint tracking field, especially for urban scenarios where real joint tracking datasets are missing. 

\section{Conclusion}
In this paper, we presented a massive CG dataset for human pose estimation and tracking which simulates realistic urban scenarios. The precise annotation of occluded joints provided by our dataset allowed us to extend a state-of-the-art network by handling occluded parts. We further integrate temporal coherency and propose a novel network capable of jointly locate people body parts and associate them across short temporal spans. Results suggest that the network, even if trained solely on synthetic data, adapts to real world scenarios when the image resolution and sharpness are high enough. We believe that the proposed dataset and architecture jointly constitute a starting point for considering tracking in surveillance as a unique process composed by detection and temporal association and can provide reliable tracklets as the input for batch optimization and re-id techniques.

\section*{Acknowledgments}
The work is supported by the Italian MIUR, Ministry of Education, Universities and Research, under the project COSMOS PRIN 2015 programme 201548C5NT. We also gratefully acknowledge the support of Panasonic Silicon Valley Lab and Facebook AI Research with the donation of the GPUs used for this research. We finally thank Marco Gianelli and Emanuele Frascaroli for developing part of the mod used to acquire JTA dataset.

\clearpage

\bibliographystyle{splncs04}
\bibliography{egbib}



\section*{Supplementary material}

# Supplementary Material

Matteo Fabbri*, Fabio Lanzi*, Simone Calderara*, Andrea Palazzi, Roberto Vezzani, and Rita Cucchiara

Department of Engineering "Enzo Ferrari"
University of Modena and Reggio Emilia, Italy
{name.surname}@unimore.it

## 1   Per Joint Results on JTA

Table 1 reports the results in term of mean average precision (mAP) per joint.

**Table 1.** Mean Average Precision (mAP) per body joint. Experiments are performed on JTA Dataset

|  | Head | Shou | Elb | Wri | Hip | Knee | Ankl | Total |
|---|---|---|---|---|---|---|---|---|
| Single Image no occ | 63.5 | 55.1 | 48.4 | 41.1 | 55.0 | 46.4 | 38.5 | 50.9 |
| Single Image + occ | 70.5 | 61.3 | 53.7 | 45.9 | 61.0 | 51.6 | 42.8 | 56.3 |
| Complete | **74.4** | **64.8** | **56.9** | **48.5** | **64.3** | **54.5** | **45.1** | **59.3** |
| [1] | 62.5 | 54.3 | 47.5 | 40.6 | 54.0 | 45.5 | 37.9 | 50.1 |

## 2   Per Sequence Results on MOT-16

Table 2 reports the metrics per sequence performed on MOT-16. A further experiment was conducted on the MOT-16 benchmark training set Figure 1 where we compared the MOTA scored by our model with and without fine-tuning on real data. Fine-tuning was performed by considering the ground truth detections and inserting a default skeleton when our Single Image model scored a false negative obtaining an automatically annotated dataset. The network was subsequently end-to-end fine-tuned, with the exception of the occlusion branch. By observing the results, we can conclude that the features extracted on our virtual world are still capable of extracting people joints in real-world images with a high resolution and sharpness (MOT16-09, MOT16-11) but with limited generalization as the image quality decreases. Nevertheless, even with a limited fine-tuning the network achieves the capability of adapting the features even in presence of a self-annotated dataset with potential errors and inaccuracies.

---

* Equal contribution.

**Table 2.** Results on MOT-16 benchmark per sequence

| Sequence | MOTA | IDF1 | MT | ML | FP | FN | IDs | FRAG |
|----------|------|------|------|------|------|------|------|------|
| MOT16-01 | 36.8 | 30.8 | 30.4 | 13.0 | 1110 | 2710 | 222 | 280 |
| MOT16-03 | 71.6 | 34.7 | 46.6 | 10.1 | 1156 | 26839 | 1723 | 2454 |
| MOT16-06 | 55.1 | 14.4 | 31.7 | 29.0 | 721 | 4159 | 302 | 323 |
| MOT16-07 | 41.9 | 29.8 | 18.5 | 16.7 | 1233 | 7759 | 489 | 713 |
| MOT16-08 | 32.5 | 22.7 | 15.9 | 34.9 | 862 | 10109 | 327 | 476 |
| MOT16-12 | 38.5 | 20.7 | 20.9 | 38.4 | 958 | 4027 | 115 | 247 |
| MOT16-14 | 16.2 | 13.0 | 4.3 | 40.2 | 3142 | 11456 | 886 | 1064 |

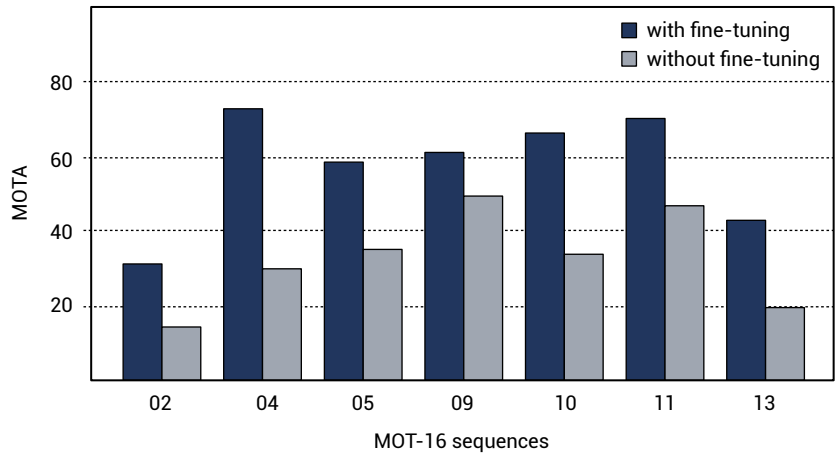

**Fig. 1.** Bar-chart of THOPA-net (trained on JTA dataset) results on MOT16 training set with and without fine-tuning on real data

## 3   JTA Dataset

Figure 2 shows information about camera distances and poses per frame of the JTA Dataset. Figure 3 and Figure 4 provide examples from the dataset exhibiting its variety in viewpoints, number of people, illuminations and scenarios.

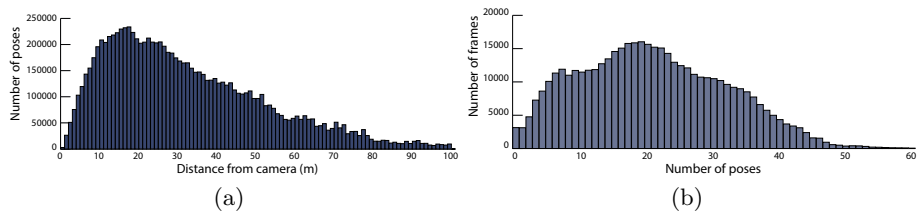

(a)                                      (b)

**Fig. 2.** Statistics about JTA Dataset. (a) Number of annotated pose per camera distance (in meters). (b) Number of frames vs. the number of annotated poses per frame

**Fig. 3.** Some images taken from JTA dataset

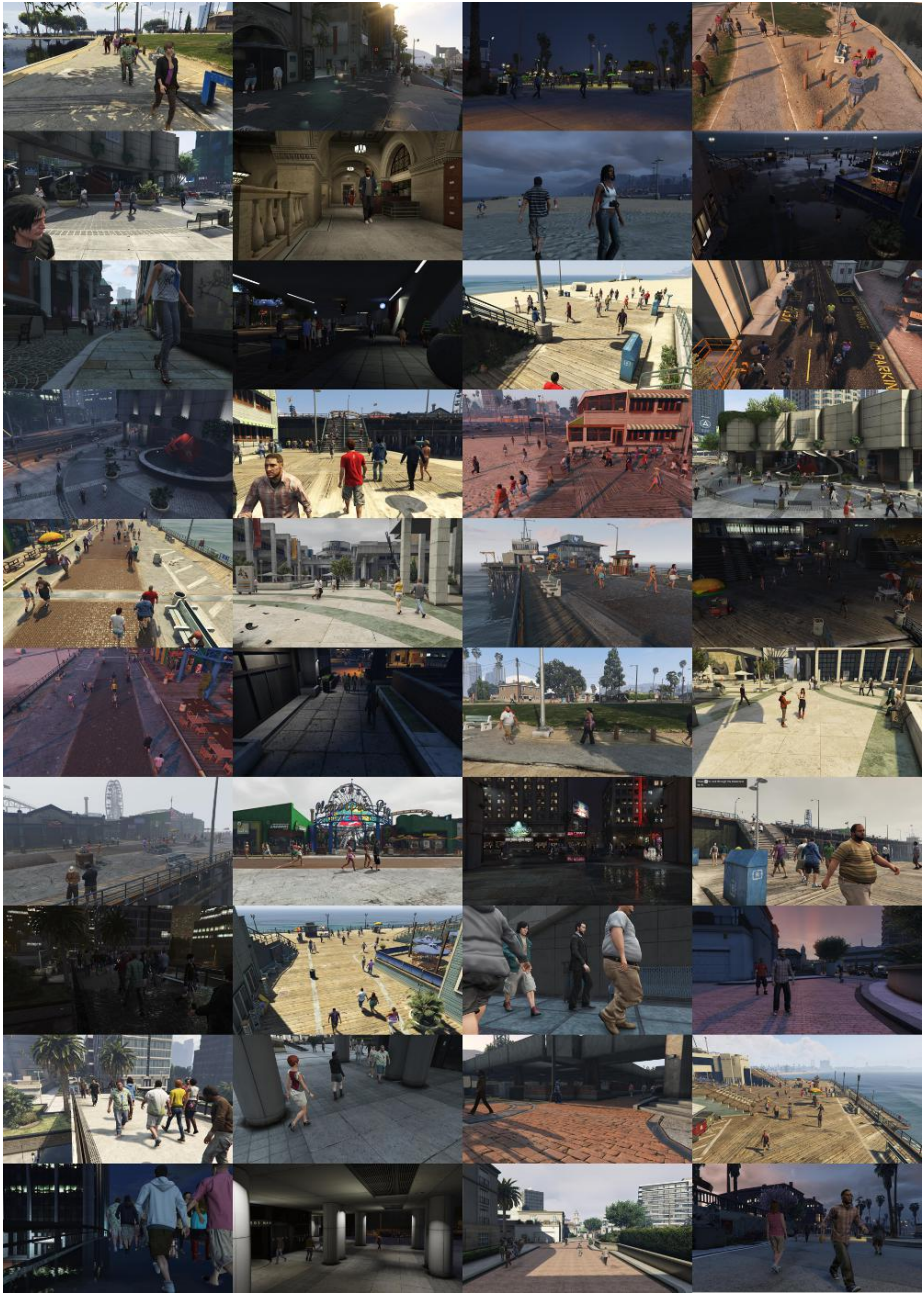

**Fig. 4.** Some images taken from JTA dataset



\end{document}